\documentclass{article}  % Comment this line out if you need a4paper

\usepackage{arxiv}

\usepackage[utf8]{inputenc} % allow utf-8 input
\usepackage[T1]{fontenc}    % use 8-bit T1 fonts
\usepackage{url}            % simple URL typesetting
\usepackage{amsfonts}       % blackboard math symbols
\usepackage{nicefrac}       % compact symbols for 1/2, etc.
\usepackage{microtype}      % microtypography
\usepackage{lipsum}

\usepackage{graphics} % for pdf, bitmapped graphics files
\usepackage{epsfig} % for postscript graphics files
\usepackage{mathptmx} % assumes new font selection scheme installed
\usepackage{times} % assumes new font selection scheme installed
\usepackage{amsmath} % assumes amsmath package installed
\usepackage{amssymb}  % assumes amsmath package installed
\usepackage{booktabs}
\usepackage{siunitx}
\usepackage{multirow}
\usepackage{subfigure}
\usepackage[dvipsnames]{xcolor}
\usepackage{hyperref}
\usepackage{comment}
\usepackage{graphicx}
\usepackage[export]{adjustbox}

\title{
Autonomous Navigation in Rows of Trees and High Crops with Deep Semantic Segmentation}

\author{Alessandro Navone$^{1}$, Mauro Martini$^{1}$, Andrea Ostuni$^{1}$, Simone Angarano$^{1}$ and Marcello Chiaberge$^{1}$ %<-this % stops a space
\thanks{$^{1}$ Department of Electronics and Telecommunications, Politecnico di Torino, 10129, Torino, Italy. \tt\footnotesize \{firstname.lastname\}@polito.it}}
\author{
 Alessandro Navone \\
  Department of Electronics and Telecommunications \\
  Politecnico di Torino\\
  Torino, TO, 10129 \\
  \texttt{alessandro.navone@polito.it} \\
   \And
 Mauro Martini \\
  Department of Electronics and Telecommunications \\
  Politecnico di Torino\\
  Torino, TO, 10129 \\
  \texttt{mauro.martini@polito.it} \\
  \And
 Andrea Ostuni \\
  Department of Electronics and Telecommunications \\
  Politecnico di Torino\\
  Torino, TO, 10129 \\
  \texttt{andrea.ostuni@polito.it} \\
  \And
 Simone Angarano \\
  Department of Electronics and Telecommunications \\
  Politecnico di Torino\\
  Torino, TO, 10129 \\
  \texttt{simone.angarano@polito.it} \\
  \And
 Marcello Chiaberge \\
  Department of Electronics and Telecommunications \\
  Politecnico di Torino\\
  Torino, TO, 10129 \\
  \texttt{marcello.chiaberge@polito.it} \\
}

\begin{document}

\maketitle
\thispagestyle{empty}
\pagestyle{empty}

%%%%%%%%%%%%%%%%%%%%%%%%%%%%%%%%%%%%%%%%%%%%%%%%%%%%%%%%%%%%%%%%%%%%%%%%%%%%%%%%
\begin{abstract}
Segmentation-based autonomous navigation has recently been proposed as a promising methodology to guide robotic platforms through crop rows without requiring precise GPS localization. However, existing methods are limited to scenarios where the centre of the row can be identified thanks to the sharp distinction between the plants and the sky. However, GPS signal obstruction mainly occurs in the case of tall, dense vegetation, such as high tree rows and orchards. In this work, we extend the segmentation-based robotic guidance to those scenarios where canopies and branches occlude the sky and hinder the usage of GPS and previous methods, increasing the overall robustness and adaptability of the control algorithm. Extensive experimentation on several realistic simulated tree fields and vineyards demonstrates the competitive advantages of the proposed solution.

\end{abstract}


%%%%%%%%%%%%%%%%%%%%%%%%%%%%%%%%%%%%%%%%%%%%%%%%%%%%%%%%%%%%%%%%%%%%%%%%%%%%%%%%
\section{Introduction}
In recent years, precision agriculture has pushed the boundaries of technology to optimize crop production, improve the efficiency of farming operations, and reduce waste \cite{zhai2020decision}. Modern farming systems must be able to extract synthetic key information from the environment, take or suggest optimal decisions based on that information, and execute them with high precision and timing. Deep learning techniques have shown great potential in realizing these systems by analyzing data from multiple sources, allowing for large-scale, high-resolution monitoring, and providing detailed insights for both human and robotic agents. The most recent advancements in deep learning also provide competitive advantages for real-world applications, such as model optimization for fast inference on low-power embedded hardware \cite{mazzia2020real, angarano2021ultra} and generalization to unseen data \cite{martini2021domain, angarano2022back, angarano2023domain}. At the same time, progress in service robotics has enabled autonomous mobile agents to embody AI perception systems and work in synergy with them to accomplish complex tasks in unstructured environments \cite{eirale2022marvin}.

\begin{figure}[t]
    \centering
    \includegraphics[width=0.8\columnwidth]{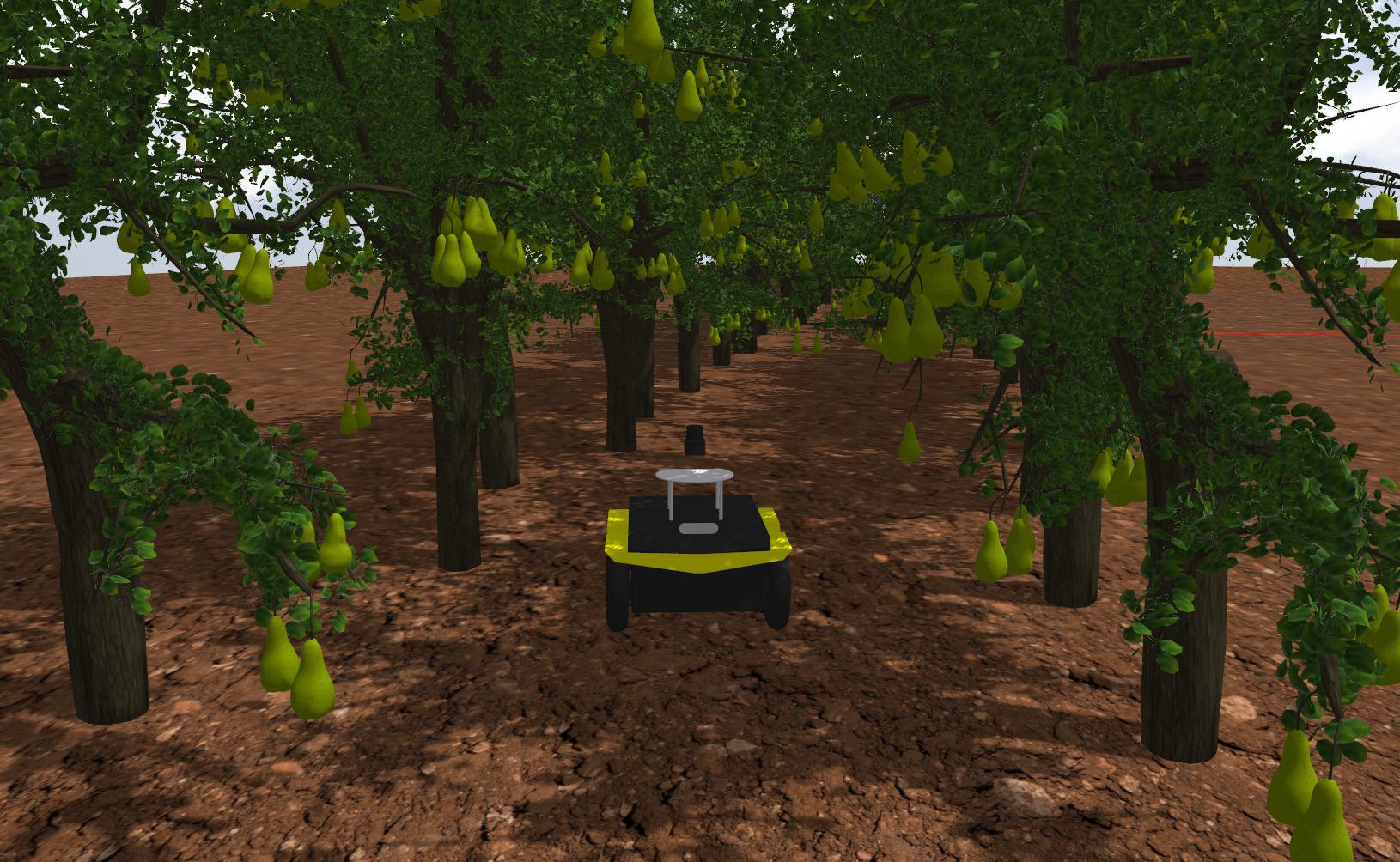}
    \caption{The proposed SegMin and SegMinD algorithms allow to precisely guide an autonomous mobile robot through a dense tree row solely using an RGB-D camera. A pear crop row in Gazebo is shown in the picture.}
    \label{fig:jackal}
\end{figure}

In particular, row-based crops are among the most studied applications (they constitute more than 75\% of all planted acres of cropland across the USA \cite{Bigelow:263079}). In this scenario, research spans localization\cite{winterhalter2021localization}, path planning \cite{salvetti2023waypoint}, navigation \cite{martini2022}, monitoring\cite{comba20192d}, harvesting \cite{harvesting}, spraying \cite{deshmukh2021design, spraygrape}, and vegetative assessment \cite{zhang2020assessment, feng2020yield}. A particularly challenging situation occurs when standard localization methods, like GPS, fail to reach the desired precision due to unfavorable weather conditions or line-of-sight obstruction. That is the case, for example, of dense tree canopies, as shown in a simulated pear orchard in Figure \ref{fig:jackal}.

Previous works have proposed position-agnostic vision-based navigation algorithms for row-based crops. A first vision-based approach was proposed in \cite{sharifi2015novel} using mean-shift clustering and the Hough transform to segment RGB images and generate the optimal central path. Later, \cite{RADCLIFFE2018165} achieved promising results using multispectral images and simply thresholding and filtering on the green channel. Recently, deep-learning approaches have been successfully applied to the task. \cite{huang2021endtoend} proposed a classification-based approach in which a model predicts the discrete action to perform. In contrast, \cite{aghi2021deep} proposed combining a segmentation model and a proportional controller to align the robot to the center of the row. Finally, a different approach was tested in \cite{martini2022} with an end-to-end controller based on deep reinforcement learning.
Although these systems proved effective in their testing scenarios, they have only been applied in simple crops where a full view of the sky favors both GPS receivers \cite{GPS_accuracy} and vision-based algorithms \cite{zaman2019cost}.

This work tackles a more challenging scenario in which dense canopies partially or totally cover the sky, and the GPS signal is very weak. We design a navigation algorithm based on semantic segmentation that exploits visual perception to estimate the center of the crop row and align the robot trajectory to it. The segmentation masks are predicted by a deep learning model designed for real-time efficiency and trained on realistic synthetic images. The proposed navigation algorithm improves on previous works being adaptive to different terrains and crops, including dense canopies. We conduct extensive experimentation in simulated environments for multiple crops. We compare our solution with previous state-of-the-art methodologies, demonstrating that the proposed navigation system is effective and adaptive to numerous scenarios.

\begin{figure*}[ht]
    \centering
    \includegraphics[width=\textwidth]{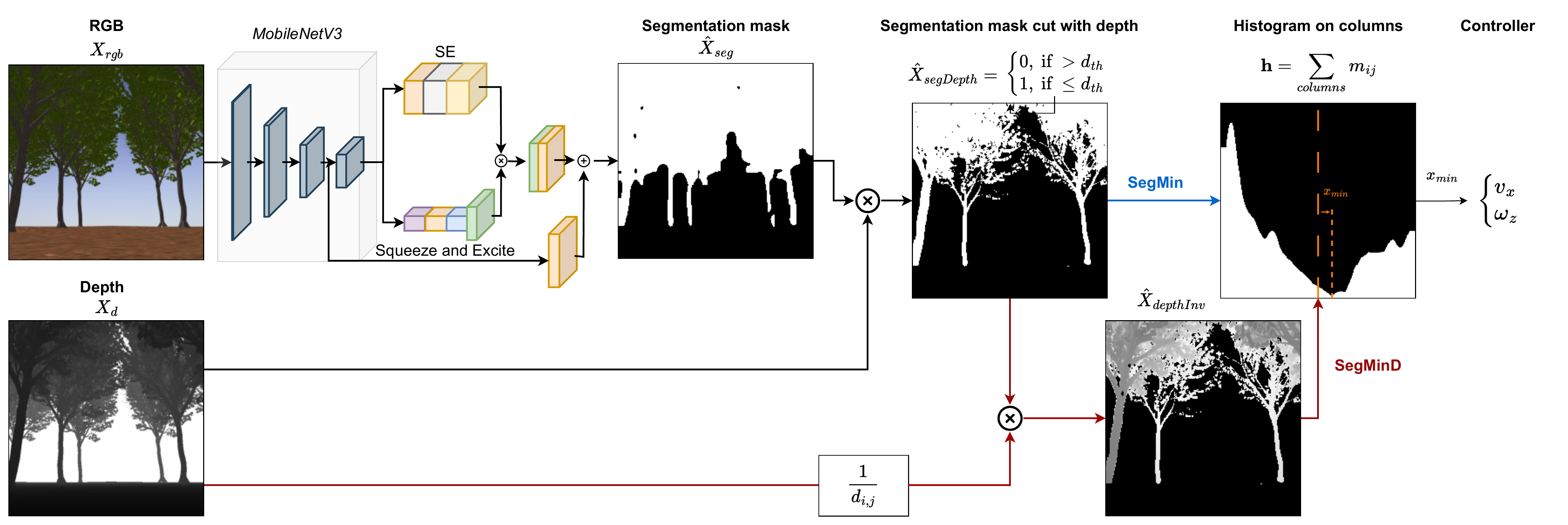}
    \caption{Scheme of the overall proposed navigation pipeline. The RGB image is fed into the segmentation network, thus the predicted segmentation mask $\hat{\textbf{X}}_{seg}^{t}$ is refined using the depth frame to obtain $\hat{\textbf{X}}_{segDepth}^{t}$. The \textcolor{RoyalBlue}{blue} arrow refers to the SegMin variant, and \textcolor{BrickRed}{red} arrows refer to the SegMinD variant to compute the sum histogram over the mask columns. Images are taken from navigation in the tall trees simulation world.}
    \label{fig:schema}
\end{figure*}

The main contributions of this work can be summarized as follows:
\begin{itemize}
    \item we present two variants of a novel approach for segmentation-based autonomous navigation in tall crops, designed to tackle challenging and previously uncovered scenarios;
    \item we test the resulting guidance algorithm on previously unseen plant rows scenarios such as high trees and pergola vineyards.
    \item we compare the new method with state-of-the-art solutions on straight and curved vineyards, demonstrating an enhanced general and robust behaviour.
\end{itemize}

The next sections are organized as follows: Section \ref{sec:methodology} presents the proposed deep-learning-based control system for vision-based position-agnostic autonomous navigation in row-based crops, from the segmentation model to the controller. Section \ref{sec:results} describes the experimental setting and reports the main results for validating the proposed solution divided by sub-system. Finally, Section \ref{sec:conclusion} draws conclusive comments on the work and suggests interesting future directions.

\section{Methodology}
\label{sec:methodology}
This work proposes a real-time control algorithm with two variants to navigate high-vegetation orchards and arboriculture fields and improve the approach presented in \cite{aghi2021deep}. The proposed system avoids exploiting the GPS signal, which can lack accuracy due to signal reflection and mitigation due to vegetation.  

The working principle of the proposed control algorithms is straightforward and exploits only the RGB-D data. Both the proposed solutions consist of four main steps:
\begin{enumerate}
    \item \label{enum:1} Semantic segmentation of the input RGB frame.   
    \item \label{enum:2} Processing of the output segmentation mask using depth frame data.
    \item \label{enum:3} Searching for the direction which leads the mobile platform towards the end of the row.   
    \item \label{enum:4} Generating linear and angular velocity commands to input the mobile robot.
\end{enumerate}

Nonetheless, the two proposed methods differ only for steps \ref{enum:2} and \ref{enum:3} in employing the depth frame data and in the generation of the path which the robot should follow. In contrast, the segmentation technique \ref{enum:1} and the command generation \ref{enum:4} are carried out similarly. A schematic representation of the proposed pipeline is described in Figure \ref{fig:schema}.

As in \cite{aghi2021deep} a first step, an RGB frame $\mathbf{X}_{rgb}^{t} \in \mathbb{R}^{h\times w \times c}$ and a depth map $\mathbf{X}_{d}^{t} \in \mathbb{R}^{h \times w}$ are acquired by a camera placed on the front of the mobile platform at each instant $t$, where $h$ and $w$ are the width of the frame and $c$ is the number of channels. The received RGB data is then fed to a segmentation neural network model $H_{seg}$, which outputs a binary segmentation mask bringing the semantic information of the input frame.
\begin{equation}
    \hat{\textbf{X}}_{seg}^{t} = H \left(\textbf{X}_{rgb}^{t}\right)
    \label{eq:segmentation}
\end{equation}
where $\hat{\textbf{X}}_{seg}^{t}$ is the estimated segmentation mask.
Moreover, the segmentation masks of the last $N$ time instants $\{t-N,\dots, t\}$ are fused to obtain more robust information.
\begin{equation}
    \hat{\textbf{X}}_{CumSeg}^{t} = \bigcup_{j = t-N}^{t} \hat{\textbf{X}}_{seg}^{t}
    \label{eq:cumulative-mask}
\end{equation}
where $\hat{\textbf{X}}_{CumSeg}^{t}$ is the cumulative segmentation mask and the operator $\bigcup$ represent the logical bitwise $OR$ operation over the last $N$ binary frames. 

Additionally, the depth map $\mathbf{X}_{d}^{t}$ is now used to consider the segmented regions between the camera position and a given depth threshold $d_{th}$ to remove useless information given by far vegetation, which is irrelevant to control the robot's movement.
\begin{equation}
    \hat{\textbf{X}}_{segDepth\substack{i = 0,\dots,h\\ j = 0,\dots,w}}^{t} (i, j)
    = 
    \begin{cases}
        0, \textrm{ if } \hat{\textbf{X}}_{CumSeg (i, j)}^{t} > d_{th}\\
        1, \textrm{ if } \hat{\textbf{X}}_{CumSeg (i, j)}^{t}\leq d_{th}
    \end{cases}
    \label{eq:depth}
\end{equation}

\noindent where $\hat{\textbf{X}}_{segDepth}$ is the resulting intersection between the cumulative segmentation frame and the depth map cut at a distance threshold $d_{th}$.

Henceforth the proposed algorithm forks in two variants, \textit{SegMin} and \textit{SegMinD}, respectively described in \ref{subsec:varA} and \ref{subsec:varB}.

\subsection{SegMin}
\label{subsec:varA}
The first variant improves the approach proposed in \cite{aghi2021deep}. After processing the segmentation mask, a sum over the column is performed to obtain a histogram $\textbf{h} \in \mathbb{R}^{w}$, quantifying how much vegetation is present on each column. Hereafter, a moving average on a window of $n$ elements is performed over the array to smooth the values and make the control more robust to punctual noise derived from the previous passages. Ideally, the minimum of this histogram $x_{h}$ corresponds to the regions where less vegetation is present and, therefore, identifies the desired central path inside the crop row. If more global minimum points are present (i.e., there is a region where no vegetation is detected), the mean of the considered points is considered to be the global minimum and, in consequence, the continuation of the row. 

\subsection{SegMinD}

\label{subsec:varB}
The second proposed approach consists of a variant of the previous algorithm, devised for wide rows with tall and thick canopies, which in the previous case would generate an ambiguous global minimum due to the constant presence of vegetation above the robot. This variant multiplies the previously processed segmentation mask for the normalized inverted depth datum. 
\begin{equation}
     \hat{\textbf{X}}_{depthInv}^{t} = \hat{\textbf{X}}_{segDepth}^{t}  \bigcap \left(1 - \dfrac{\textbf{X}_{d}^t}{d_{th}}\right)
     \label{eq:depth-inv}
\end{equation}

where $\bigcap$ represents the element-wise multiplication between the binary mask $\hat{\textbf{X}}_{segDepth}^{t}$ and the inverted depth frame $\hat{\textbf{X}}_{segDepth}^{t}$ normalized over the depth threshold $d_{th}$. As in the previous case, the sum over the column is performed to obtain the 1D array $\textbf{h}$ and, later on, the smoothing through a moving average. The introduced modification allows the closer elements to exert a greater influence on identifying the row direction.

\subsection{Segmentation Network}
We adopt the same network used in previous works on real-time crop segmentation \cite{aghi2021deep, angarano2023domain}. The model consists of a MobilenetV3 backbone for feature extraction and an efficient LR-ASPP segmentation head \cite{howard2019searching}. In particular, the LR-ASPP leverages effective modules such as depth-wise convolutions, channel-wise attention, and residual skip connections to provide an effective trade-off between accuracy and inference speed. The model is trained with a similar procedure to \cite{angarano2023domain} on the AgriSeg dataset\footnote{\url{https://pic4ser.polito.it/AgriSeg}}. Further details on the training strategy and hyperparameters are provided in Section \ref{sec:results}.

\subsection{Robot heading control}
The objective of the controller pipeline consists in keeping the mobile platform at the center of the row, which, in this work, is considered equivalent to keeping the row center in the middle of the camera frame. Therefore, as defined in the previous step, the minimum of the histogram should be centered in the frame width. The distance $d$ from the center of the frame and the minimum is defined as:
\begin{equation}
    d = x_{h} - \dfrac{w}{2}
\end{equation}

The linear and angular velocities are then generated through custom functions as in \cite{cerrato2021deeplearning}.
\begin{equation}
    v_{x} = v_{x, max} \left(1 - \frac{d^{2}}{\frac{w}{2}^{2}} \right)
\end{equation}

\begin{equation}
    \omega_{z} = -\omega_{z, gain}\cdot d
    \label{eq:ang_vel}
\end{equation}

where $v_{x, max}$ is the maximum achievable linear speed and  $\omega_{z, gain}$ is the andular gain. In order to avoid abrupt changes in the robot's motion, the final velocities $\bar{v}_{x}$ and $\bar{\omega}_{z}$ commands are smoothed with an Exponential Moving Average (EMA) as:
\begin{equation}
    \begin{bmatrix}
        \bar{v}_{x}^{t} \\ \bar{\omega}_{z}^{t}
    \end{bmatrix}
     = (1 - \lambda) \begin{bmatrix}
        \bar{v}_{x}^{t-1} \\ \bar{\omega}_{z}^{t-1}
    \end{bmatrix}
    + \lambda \begin{bmatrix}
        {v}_{x}^{t} \\ \omega_{z}^{t}
    \end{bmatrix}
\end{equation}

where $t$ is the time step and $\lambda$ is a chosen weight.

\section{Experiments and Results}
\label{sec:results}

\subsection{Simulation Environment}
The proposed control algorithm was tested through the use of Gazebo\footnote{\url{https://gazebosim.org}} simulation software. The software was selected because of its compatibility with ROS 2 and can incorporate plugins that simulate sensors, such as cameras. A Clearpath Jackal model was utilized to assess the algorithm's effectiveness. The URDF file, available through Clearpath Robotics, contains all the necessary information regarding the mechanical structure and joints of the robot. During the simulation, an Intel Realsense D435i plugin was utilized, positioned 20 cm in front of the robot's center, and tilted $15^{\circ}$ upwards. This positioning gave the camera a better view of the upper branches of trees. 

The navigation algorithm was tested in four different custom simulation environments: a common vineyard, a pergola vineyard characterized by vine poles and shoots above the row, a pear field constituted by small size trees, and a high trees field where canopies of the trees are merged above the row. Each simulated field adopts a different terrain, miming the irregularity of uneven terrain. The detailed measurements of the simulation world are described in Table \ref{tab:mondi}.

During the experimental part of this work, we consider frame dimensions equal to $(h, w) = (224, 224)$, which is the same size as the input and the output of the neural network model, with the number of channels $c = 3$. The maximum linear velocity has been fixed to $v_{x, max} = 0.5 m/s$, and the maximum angular velocity has been fixed to $\omega_{z, max} = 1 rad/s$. The angular velocity gain $\omega_{z, gain}$ has been fixed to $0.01$ and the EMA buffer size has been fixed to 3. The depth threshold has been changed according to the various crops. In particular, it has been fixed to 5 m in the case of vineyards, while it was increased to 8 m for pear trees and pergola vineyards and 10 m for tall trees. 

\begin{figure}[ht]
\setlength{\fboxsep}{0pt}
\centering
\resizebox{0.9\columnwidth}{!}{%
\begin{tabular}{cccc}
& RGB & Masked Depth & Histogram\\ 
\vspace{5pt}
(a) & \framebox{\includegraphics[width=.25\columnwidth, valign=m]{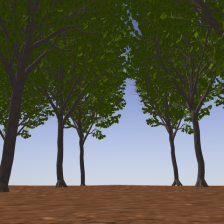}} & \framebox{\includegraphics[width=.25\columnwidth, valign=m]{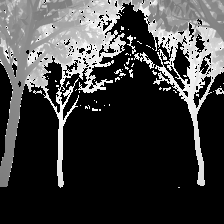}} & \framebox{\includegraphics[width=.25\columnwidth, valign=m]{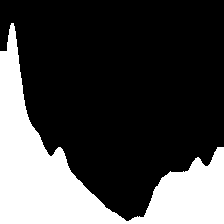}}\\ 
\vspace{5pt}
(b) & \framebox{\includegraphics[width=.25\columnwidth, valign=m]{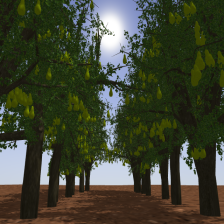}} & \framebox{\includegraphics[width=.25\columnwidth, valign=m]{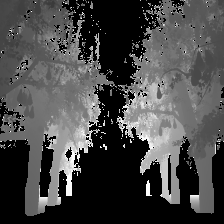}} & \framebox{\includegraphics[width=.25\columnwidth, valign=m]{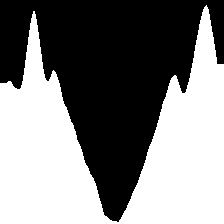}}\\
\vspace{5pt}
(c) & \framebox{\includegraphics[width=.25\columnwidth, valign=m]{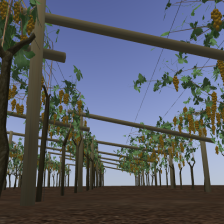}} & \framebox{\includegraphics[width=.25\columnwidth, valign=m]{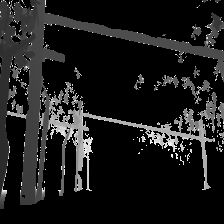}} & \framebox{\includegraphics[width=.25\columnwidth, valign=m]{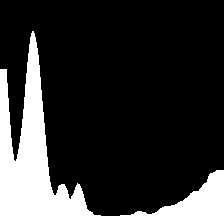}}\\
\vspace{5pt}
    (d) & \framebox{\includegraphics[width=.25\columnwidth, valign=m]{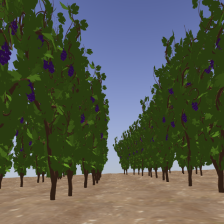}} & \framebox{\includegraphics[width=.25\columnwidth, valign=m]{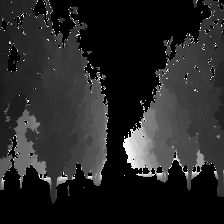}} & \framebox{\includegraphics[width=.25\columnwidth, valign=m]{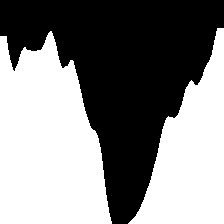}}\\
\end{tabular}
}
\caption{Sample outputs of the proposed SegMinD algorithm for High Trees (a), Pear Trees (b), Pergola Vineyard (c), and Vineyard (d). Predicted segmentation masks are refined cutting values exceeding a depth threshold. The sum over mask columns provide the histograms used to identify the centre of the row as its global minimum.}
\end{figure}
\begin{table}[t]
\centering
\caption{Size of the different simulated crops, referring to the average values of the distance between rows, the distance between plants on the row and the heights of the plants.}
%\resizebox{0.98\columnwidth}{!}{%
\begin{tabular}{lccc}
\toprule
\textbf{Gazebo worlds}    & \textbf{Rows distance {[}m{]}} & \textbf{Plant distance  {[}m{]}} & \textbf{Height {[}m{]}} \\ \midrule
\textbf{Common vineyard}  & 1.8                         & 1.3                             & 2.0               \\
\textbf{Pergola vineyard} & 6.0                           & 1.5                             & 2.9             \\
\textbf{Pear field}       & 2.0                           & 1.0                               & 2.9            \\
\textbf{High trees field} & 7.0                           & 5.0                               & 12.5            \\ \bottomrule
\end{tabular}%
%}

\label{tab:mondi}
\end{table}

\begin{figure}[ht]
    \centering
    \setlength{\fboxsep}{0pt}
    \centering
    \resizebox{0.9\columnwidth}{!}{%
    \begin{tabular}{c c}
    \includegraphics[width=0.5\columnwidth]{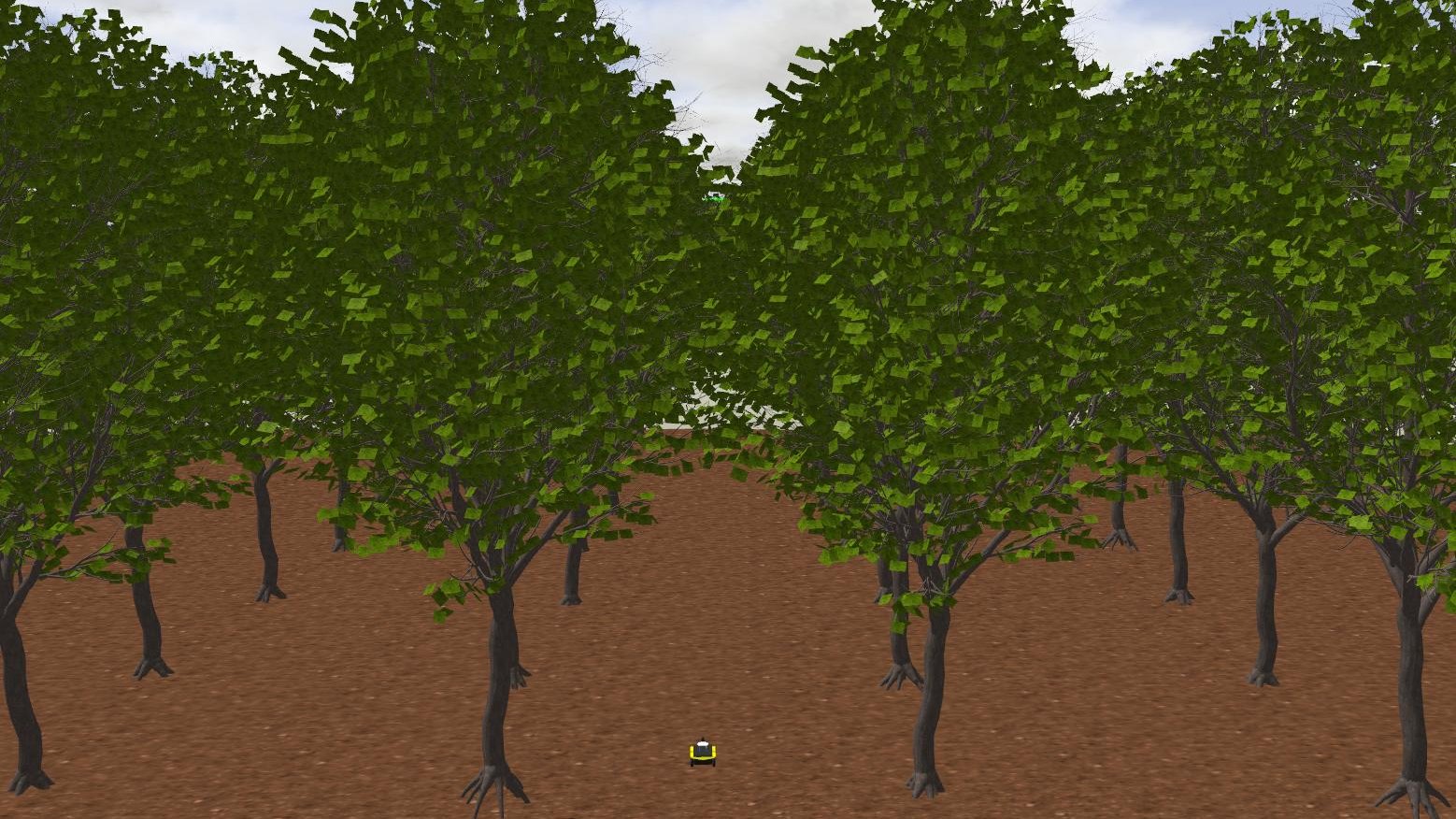} & \includegraphics[width=0.5\columnwidth]{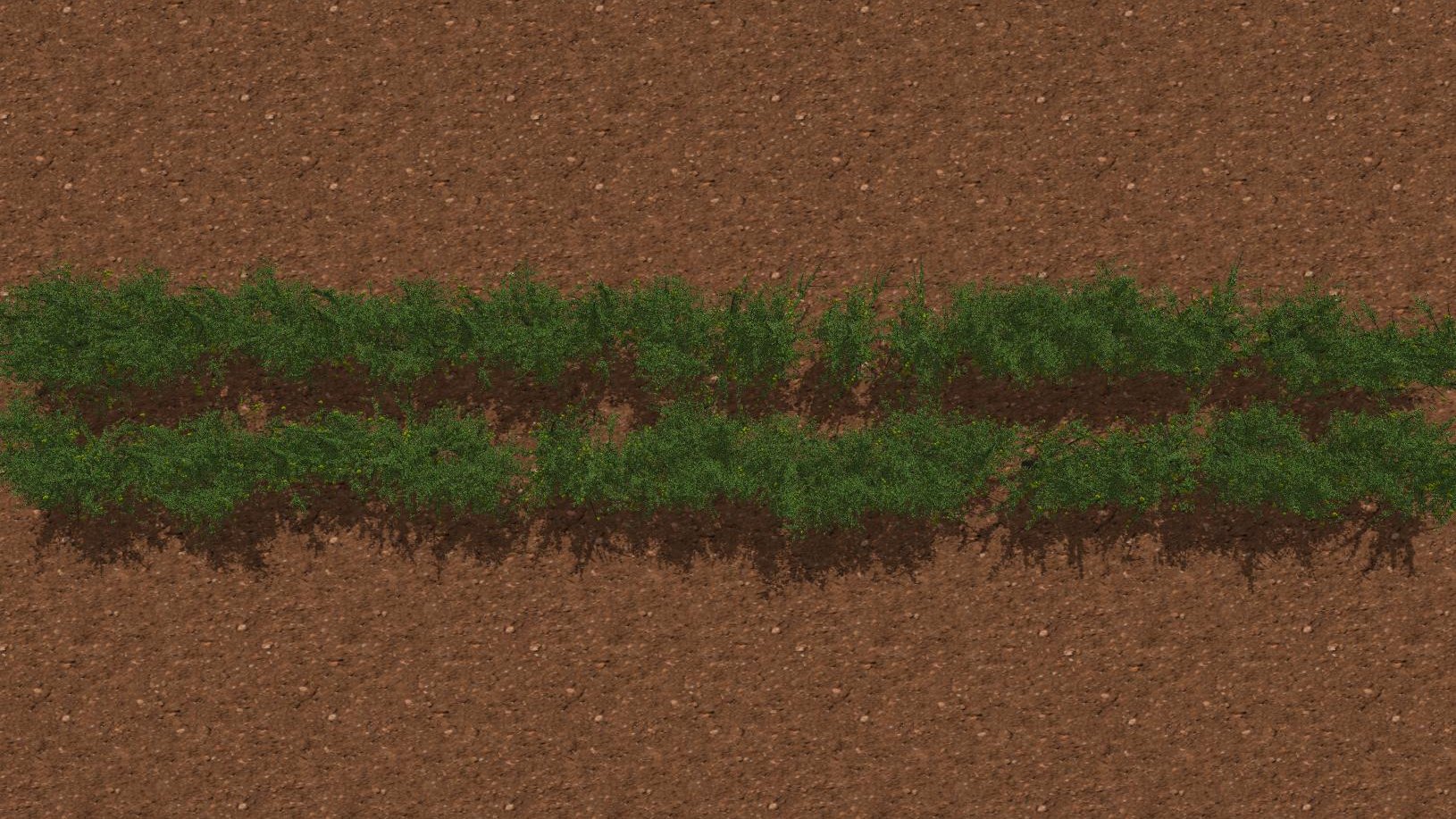} \\
    (a) & (b)\\
    \includegraphics[width=0.5\columnwidth]{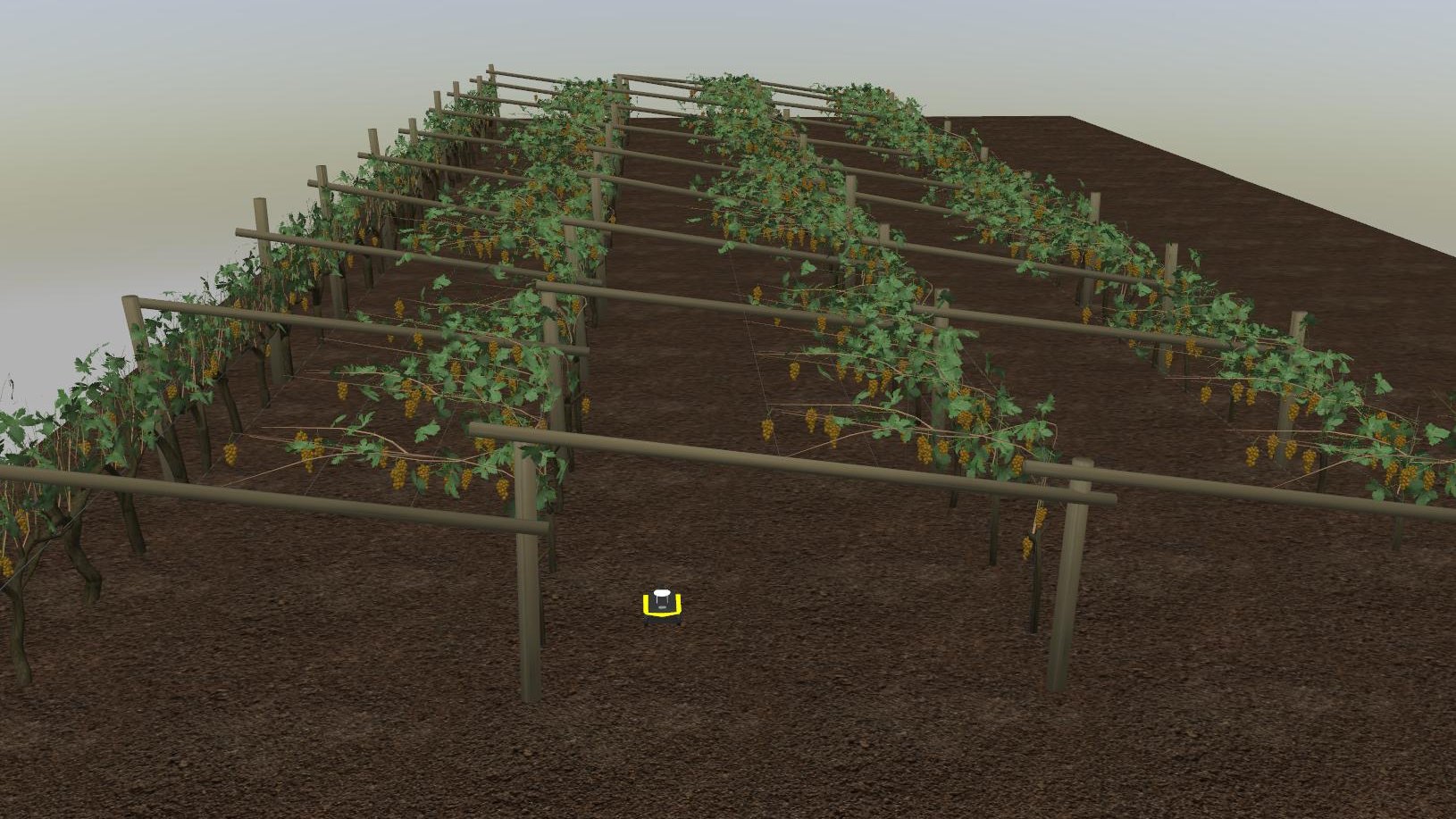} & \includegraphics[width=0.5\columnwidth]{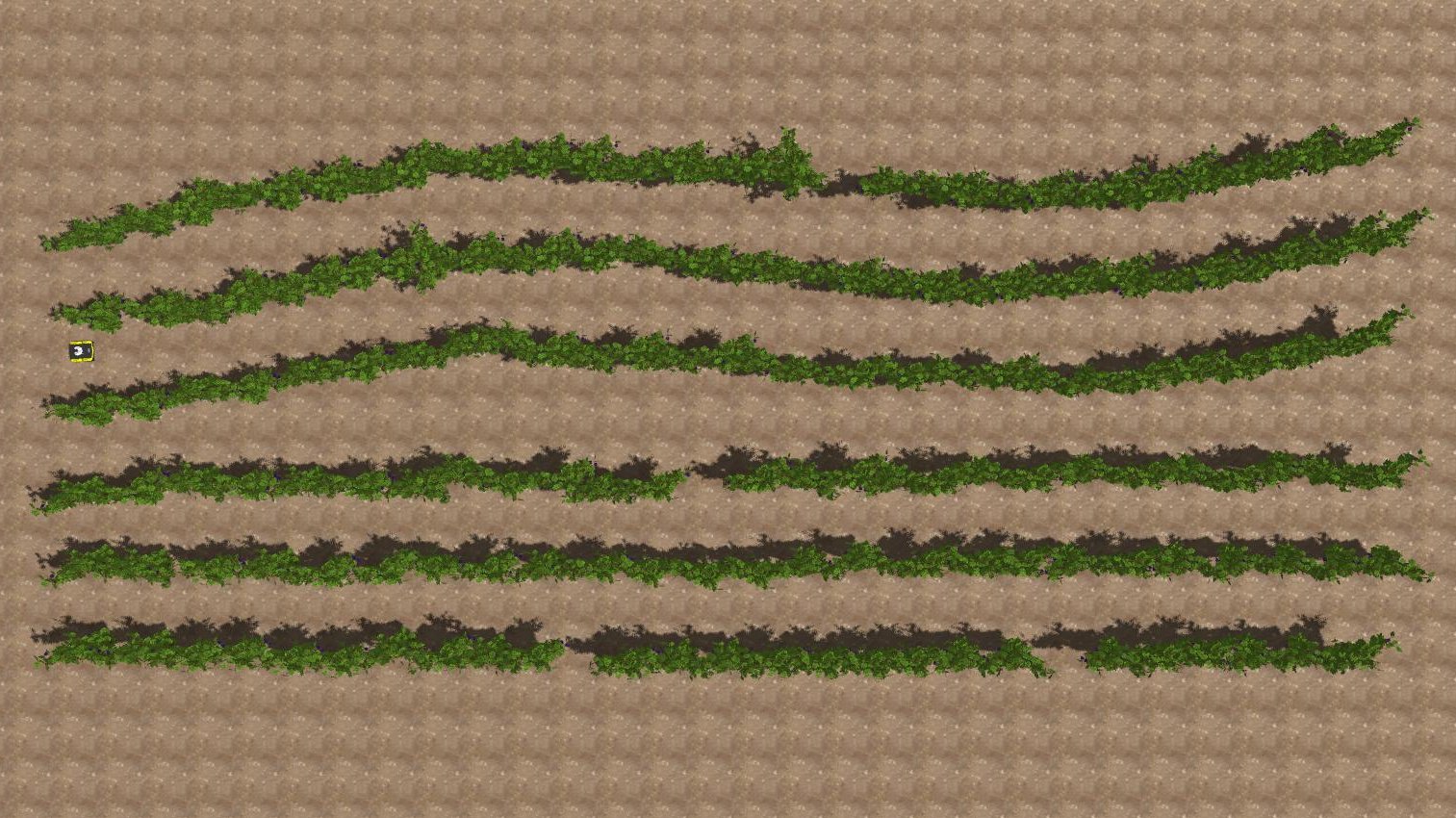} \\
    (c) & (d)\\
    \end{tabular}
    }
    \caption{Gazebo simulated environments used to test the SegMin approach in relevant different crops rows: wide rows composed of high trees (a), a narrow pear tree row (b), a pergola vineyard with asymmetric rows (c), straight and curved vineyard rows (d). In the last case, the tests were carried out in the second row from above and the second row from below.}
    \label{fig:mondi}
\end{figure}

\subsection{Segmentation Network Training and Evaluation} % SIMO OK
We train the crop segmentation model using a subset of the AgriSeg segmentation dataset \cite{angarano2023domain}. In particular, for the High Tree and Pear crops, we train on Generic Tree splits 1 and 2, and on Pear; for Vineyards, we train on Vineyard and Pergola Vineyard (note that the testing environments are different from the ones from which the training samples are generated). In both cases, the model is trained for 50 epochs with Adam optimizer and learning rate $3\times10^{-4}$. We apply data augmentation by randomly applying cropping, flipping, greyscaling, and random jitter to the images. Our experimentation code is developed in Python 3 using TensorFlow as the deep learning framework. We train models starting from ImageNet pretrained weights, so the input size is fixed to (224 × 224). The All the training runs are performed on a single Nvidia RTX 3090 graphic card.

\subsection{Navigation Results} 

\begin{table*}[t]
\centering
\caption{Navigation results obtained in different test fields with the SegMin, SegMinD and previous work SegZeros segmentation-based algorithms. The metrics test the effectiveness of the navigation (clearance time) and its precision with Mean Absolute Error (MAE) and Mean Squared Error (MSE) between obtained and ground truth path. The cumulative heading average $\gamma [rad]$, the mean linear velocity $v_{avg} [m/s]$ and the standard deviation of the angular velocity $\omega_{stddev} [rad/s]$ commands provide relevant kinematic information of the robot while navigating.}
\label{tab:test_results}
\resizebox{\columnwidth}{!}{%
\begin{tabular}{@{}llccccccc@{}}
\toprule
\textbf{Test Field} &
  \textbf{Method} &
  \multicolumn{1}{c}{\textbf{Clearance time {[}s{]}}} &
  \multicolumn{1}{c}{\textbf{MAE {[}m{]}}} &
  \multicolumn{1}{c}{\textbf{MSE {[}m{]}}} &
  \multicolumn{1}{c}{\textbf{Cum. $\pmb{\gamma_{avg}}$ [rad]}} &
  \multicolumn{1}{c}{\textbf{$\pmb{v_{avg} [m/s]}$}} &
  \multicolumn{1}{c}{\textbf{$\pmb{\omega_{std dev} [rad/s]}$}} \\ \midrule
\textbf{High Trees}        & SegMin   & \pmb{40.409 ± 0.117} & 0.265 ± 0.005 & 0.084 ± 0.003 & 0.079 ± 0.001 & 0.487 ± 0.000 & 0.054 ± 0.002 \\
                  & SegMinD  & 40.440 ± 0.515       & \pmb{0.174 ± 0.006} & \pmb{0.036 ± 0.002} & 0.048 ± 0.002 & 0.484 ± 0.006 & 0.063 ± 0.019 \\\midrule
\textbf{Pear Trees}        & SegMin   & \pmb{42.058 ± 1.228} & 0.034 ± 0.012 & \pmb{0.002 ± 0.001} & 0.013 ± 0.002 & 0.483 ± 0.003 & 0.108 ± 0.054 \\
                  & SegMinD  & 42.259 ± 1.912       & \pmb{0.031 ± 0.017} & \pmb{0.002 ± 0.002} & 0.016 ± 0.004 & 0.477 ± 0.009 & 0.026 ± 0.004 \\\midrule
\textbf{Pergola Vineyard}  & SegMin   & \pmb{40.859 ± 0.386} & \pmb{0.077 ± 0.011} & \pmb{0.011 ± 0.003} & 0.030 ± 0.022 & 0.479 ± 0.003 & 0.174 ± 0.021 \\
                  & SegMinD  & 41.135 ± 0.329       & 0.097 ± 0.052 & 0.015 ± 0.014 & 0.029 ± 0.011 & 0.475 ± 0.004 & 0.204 ± 0.032 \\\midrule
\textbf{Straight Vineyard} & SegMin   & \pmb{50.509 ± 0.305} & \pmb{0.105 ± 0.003} & \pmb{0.014 ± 0.001} & 0.033 ± 0.002 & 0.487 ± 0.000 & 0.079 ± 0.011 \\
                  & SegMinD  & 50.629 ± 0.282       & 0.110 ± 0.005 & 0.018 ± 0.003 & 0.026 ± 0.009 & 0.486 ± 0.001 & 0.088 ± 0.005 \\
                  & SegZeros & 53.695 ± 1.029       & 0.138 ± 0.025 & 0.024 ± 0.010 & 0.027 ± 0.004 & 0.457 ± 0.008 & 0.089 ± 0.008 \\\midrule
\textbf{Curved Vineyard}   & SegMin   & 53.321 ± 0.249       & 0.115 ± 0.008 & 0.017 ± 0.002 & 0.036 ± 0.008 & 0.487 ± 0.001 & 0.088 ± 0.021 \\
                  & SegMinD  & \pmb{51.444 ± 1.030} & \pmb{0.093 ± 0.005} & \pmb{0.012 ± 0.001} & 0.015 ± 0.004 & 0.484 ± 0.007 & 0.065 ± 0.008 \\
                  & SegZeros & 71.048 ± 27.132      & 0.108 ± 0.044 & 0.019 ± 0.009 & 0.045 ± 0.008 & 0.395 ± 0.127 & 0.114 ± 0.039 \\ 
                  \bottomrule
\end{tabular}%
}
\end{table*}

The overall navigation pipeline of SegMin and its variant SegMinD are tested in realistic crops fields in simulation using relevant metrics for visual-based control without precise localization of the robot, as done in previous works \cite{aghi2021deep, martini2022}. The camera frames are published at a frequency of 30 Hz, while the inference is carried out at 20 Hz, and the controllers publish the velocity commands at 5 Hz.The evaluation has been performed using the testing package of the open-source PIC4rl-gym\footnote{\url{https://github.com/PIC4SeR/PIC4rl\_gym}} in Gazebo \cite{martini2022pic4rl}. The selected metrics aims at evaluating the effectiveness of the navigation (clearance time) as well as the precision, quantitatively comparing the obtained trajectories with a ground truth one though Mean Absolute Error (MAE) and Mean Squared Error (MSE). The ground truth trajectories have been computed averaging the curve obtained interpolating the plants poses in the rows. For the asymmetric pergola vineyard case, the row is intended as the portion of the pergola without vegetation on top, as shown in Figure \ref{fig:mondi} (c). The response of the algorithms to terrain irregularity and rows geometry is also studied including in the test significant kinematic information of the robot. The cumulative heading average $\gamma [rad]$ along the path is considered, together with the mean linear velocity $v_{avg} [m/s]$ and the standard deviation of the angular velocity $\omega_{stddev} [rad/s]$ commands predicted to keep the robot correctly oriented. The mean value of $\omega$ is always close to zero due to the consecutive correction of the robot orientation.

The complete results collection is reported in Table \ref{tab:test_results}. For each metric, an average value and the standard deviation are indicated, since all the experiments have been repeated over 3 runs on a 20 m long track in each crop row. The proposed method demonstrates to solve the problem of guiding the robot through trees rows with thick canopies (high trees and pears) without a localization system, as well as in peculiar scenarios such as the pergola vineyards. The identification of plants branches and wooden supports hinder the usage of previously existing segmentation-based solutions, that were based on the assumption of finding a free passage solely considering the zeros of the binary segmentation mask \cite{aghi2021deep}. We refer to this previous method as SegZeros in the results comparison, that we tested using the same segmentation neural network.

\begin{figure}[ht]
    \setlength{\fboxsep}{0pt}
    \centering
    \resizebox{0.9\columnwidth}{!}{%
    \begin{tabular}{c c c}
         RGB & SegMin & SegMinD  \\
         \framebox{\includegraphics[width=0.3\columnwidth]{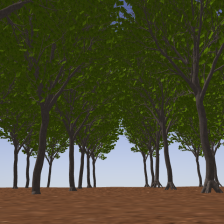}} & \framebox{\includegraphics[width=0.3\columnwidth]{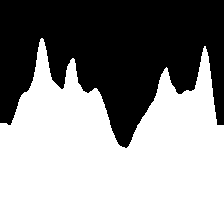}} & \framebox{\includegraphics[width=0.3\columnwidth]{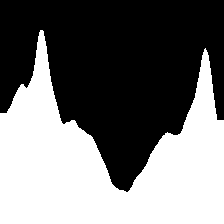}}
    \end{tabular}
    }
    \caption{Comparison of the two histograms obtained using the two different algorithms, given the RGB frame on the right. It can be noticed how SegMinD offers a narrower and less ambiguous global minimum point.}
    \label{fig:histos}
\end{figure}

The SegMin approach based on histogram minimum search demonstrate to be a robust solution to guide the robot through trees rows. The introduction of the depth inverse values as weighting function allows SegMinD to be further increase the precision of the algorithm in following the central trajectory of the row in complex cases such as wide rows (high trees) and curved rows (curved vineyard). The different sum histograms obtained with SegMin and SegMinD are directly compared in Figure \ref{fig:histos}, showing the sharper trend and the global minimum isolation obtained including the depth values. Moreover, the novel methods show competitive performance also with standard crops rows where a free passage to the end of the row can be seen in the mask without the disturbance of canopies. The histogram minimum approach significantly reduce the navigation time and the trajectory precision in vineyard rows (straight and curved) compared to previous segmentation-based baseline method. The search of plant-free zero clusters in the map results to be less robust and efficient, leading the robot to undesired stops during the navigation, and to an overall slower and more oscillating behaviour.

\begin{figure}[ht]
    \centering
    \includegraphics[width=0.9\columnwidth]{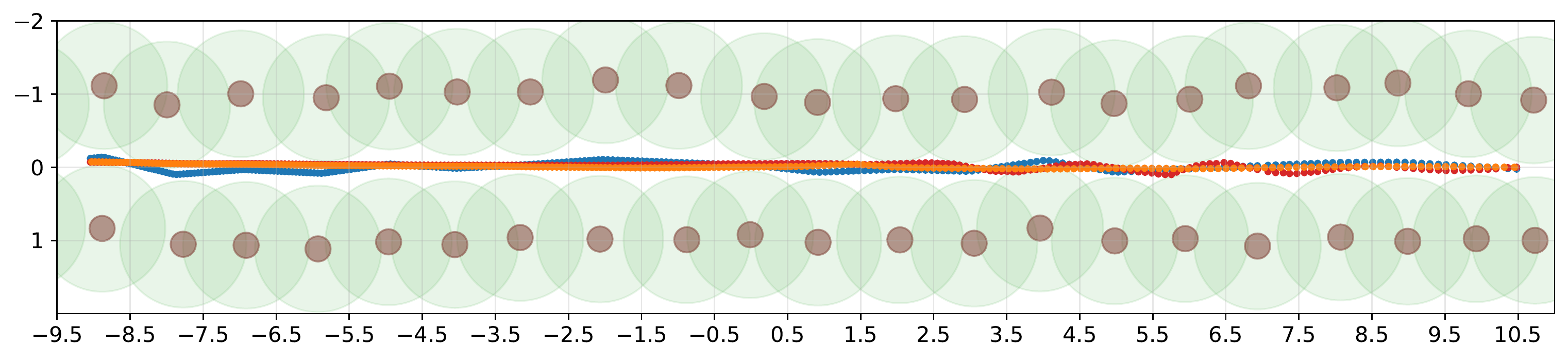}
    \includegraphics[width=0.9\columnwidth]{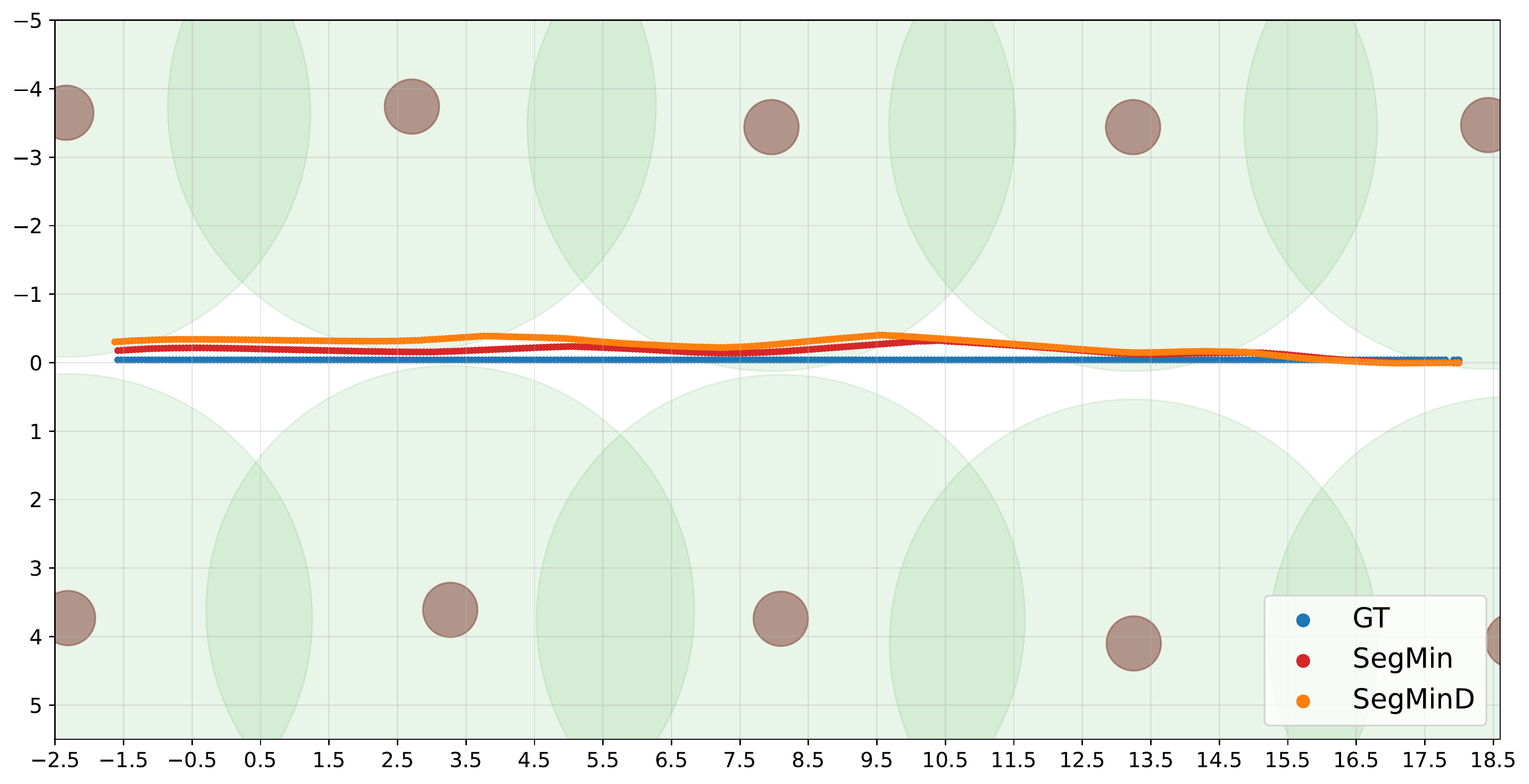}
    \includegraphics[width=0.9\columnwidth]{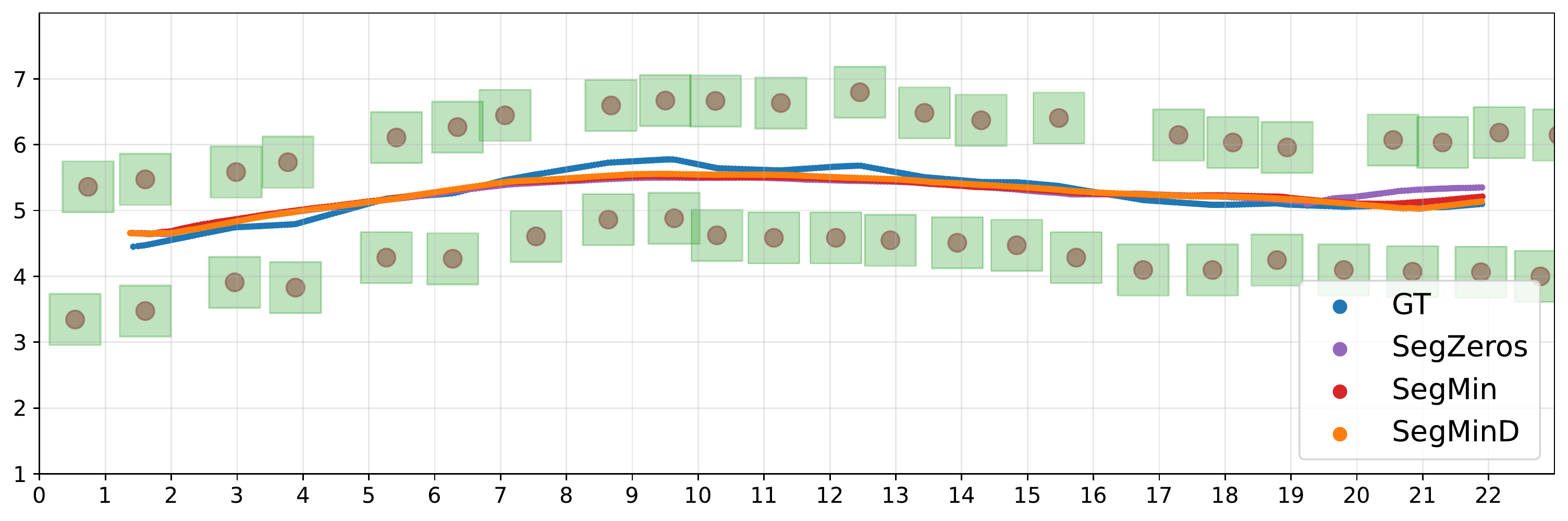}
    \caption{Trajectories comparison between our proposed algorithms (SegMin and SegMinD) and the ground truth central path (GT): Pears (top), High Trees (center), Curved Vineyard (bottom). In the last graph, the trajectory generated with the SegZeros algorithm is also reported for comparison.}
    \label{fig:trajectories}
\end{figure}

Nonetheless, the trajectories obtained with the SegMin, SegMinD and SegZeros algorithms are also visually shown in Figure \ref{fig:trajectories} inside representative scenarios: a cluttered, narrow row with small pear trees, a wide row with high trees, and curved vineyards with state-of-the-art method SegZeros.

\section{Conclusions}
\label{sec:conclusion} 
In this work, we presented a novel method to guide to a service autonomous platform through crops rows where a precise localization signal is often occluded by the vegetation. Trees rows represented an open problem in row crops navigation, since previous works based on image segmentation or processing fail due to the presence of branches and canopies covering the free passage for the rover in the image. The proposed pipeline SegMin and SegMinD overcome this limitation introducing a global minimum search on the sum histogram over the mask columns. The experiments conducted demonstrate the ability to solve the navigation task in wide and narrow trees rows and, nonetheless, the improvement in efficiency and robustness provided by our method over previous works in generic vineyards scenarios.

Future works will see the test of the overall system in real-world trees rows, orchards and vineyards to further validate the robustness of the solution with respect to sim2real gap problems and hardware resources. A 
 successful outcome is expected according to the efficient architecture adopted for the segmentation neural network, thought for low-power system applications, and the realistic features of training data and simulation. 


%%%%%%%%%%%%%%%%%%%%%%%%%%%%%%%%%%%%%%%%%%%%%%%%%%%%%%%%%%%%%%%%%%%%%%%%%%%%%%%%
\section*{Acknowledgements} 
This work has been developed with the contribution of Politecnico di Torino Interdepartmental Centre for Service Robotics PIC4SeR\footnote{\url{www.pic4ser.polito.it}}.

%%%%%%%%%%%%%%%%%%%%%%%%%%%%%%%%%%%%%%%%%%%%%%%%%%%%%%%%%%%%%%%%%%%%%%%%%%%%%%%%

\bibliographystyle{unsrt} 
\bibliography{references}
\end{document}